\documentclass[letterpaper]{article}

\usepackage{natbib}
\usepackage{graphicx}
\usepackage{arxiv}
\usepackage{url,hyperref,cleveref}
\usepackage{booktabs}

\usepackage{lipsum}

\title{Neutrally Evolving Interlocking Complexity in the Quandary Den}

\author{
    Andrew Walsh$^{1}$
    \mbox{}\\
    $^1$Independent Researcher, United States of America \\
    andywalsh1218@gmail.com
}

\begin{document}

\maketitle

\begin{abstract}
    Molecular biology features numerous complexes of proteins that coordinate in an interlocking fashion to fulfill different functions.
    Adaptive evolution explains some of this complexity, but needn't be the default when neutral explanations suffice.
    A new artificial life model ``organism,'' the Quandary Den, is introduced to explore different neutral evolution scenarios where complexity increases in the absence of greater informational needs.
    Two interlocking complexity scenarios emerge.
    Subfunctionalization leads to functionality diffusing through the complex.
    Masking allows intracomplex interference to accumulate genetically, requiring that it be blocked at the level of expression.
\end{abstract}

\section{Introduction}

Why is so much of molecular biology about protein complexes instead of protein simples (\citep{Lynch13}?
Of course, some cellular functions are accomplished by single proteins, but many are not.
Sometimes it is clear why a modular task is well-served by a multi-component solution \citep{Borowicz20, Cebecauer10}.
And if a single protein solution is not possible or not as effective, then the origin of that complexity can be understood in adaptive terms.
But, for example, when a protein complex does the same job that a single protein apparently used to do \citep{Pillai20}, an adaptive explanation may not suffice.

Explanations in terms of neutral evolution are a significant alternative.
\citet{Muller1918} proposed a generic two-step process for adding elements, resulting in complexes with ``interlocking action.'' 
Take a system that can already perform a function, add a new element which is initially not essential, and then change the original system so that they can no longer function without the added element.
The end result is a more complex way of accomplishing what could be done previously.
More recently, this notion and related ideas have been described as constructive neutral evolution (CNE) \citep{Stoltzfus99, Stoltzfus12}, resulting in ``irremediable complexity'' (\citep{Gray10}.

As described, this process involves no selection \emph{for} complexity.
It will typically involve a degree of negative selection against loss of the function that was present at the beginning, removing new changes which reduce or eliminate that function.
That also prevents the removal of added elements that have become necessary, ratcheting up the number of elements and thus the complexity \citep{Lukes11}.
Examples of mutations that prevent the subsequent loss of added elements have been observed \citep{Hochberg20, Depres24}.

Alternatively, the proposed zero-force law of evolution (ZFEL) says that complexity can increase in an even broader set of conditions, without negative selection \citep{McShea19}.
In this scenario, complexity increases not because of a ratchet but because of an asymmetry in the probability of becoming more similar and becoming more divergent.

Creating and manipulating true neutral conditions in biological contexts to compare these processes is challenging.
Establishing neutrality in the past may require the use of indirect analysis \citep{MunozGomez21}.
Artificial life experiments offer a complementary investigative option.
The type and degree of selection operating in any given experiment can be finely tuned.
And using neutral evolution baselines is already an established technique in artificial life \citep{BSBP}.

Fine control of selection and high repeatability also makes it possible to investigate whether complexity tends towards the largest volume of configuration space or whether there is a one-directional ratchet that can push a system out of that largest volume.
As long as there are multiple parts that can be added without effect, then the set of solutions with $n+1$ parts will be bigger than the set of solutions with $n$ parts since the former includes all of the latter with a neutral part added.
In that case, an increase in complexity may be the result of ergodic exploration of configuration space rather than ratcheting.

\section{Quandary Den}

\begin{figure}
    \centering
    \includegraphics[width=.4\textwidth]{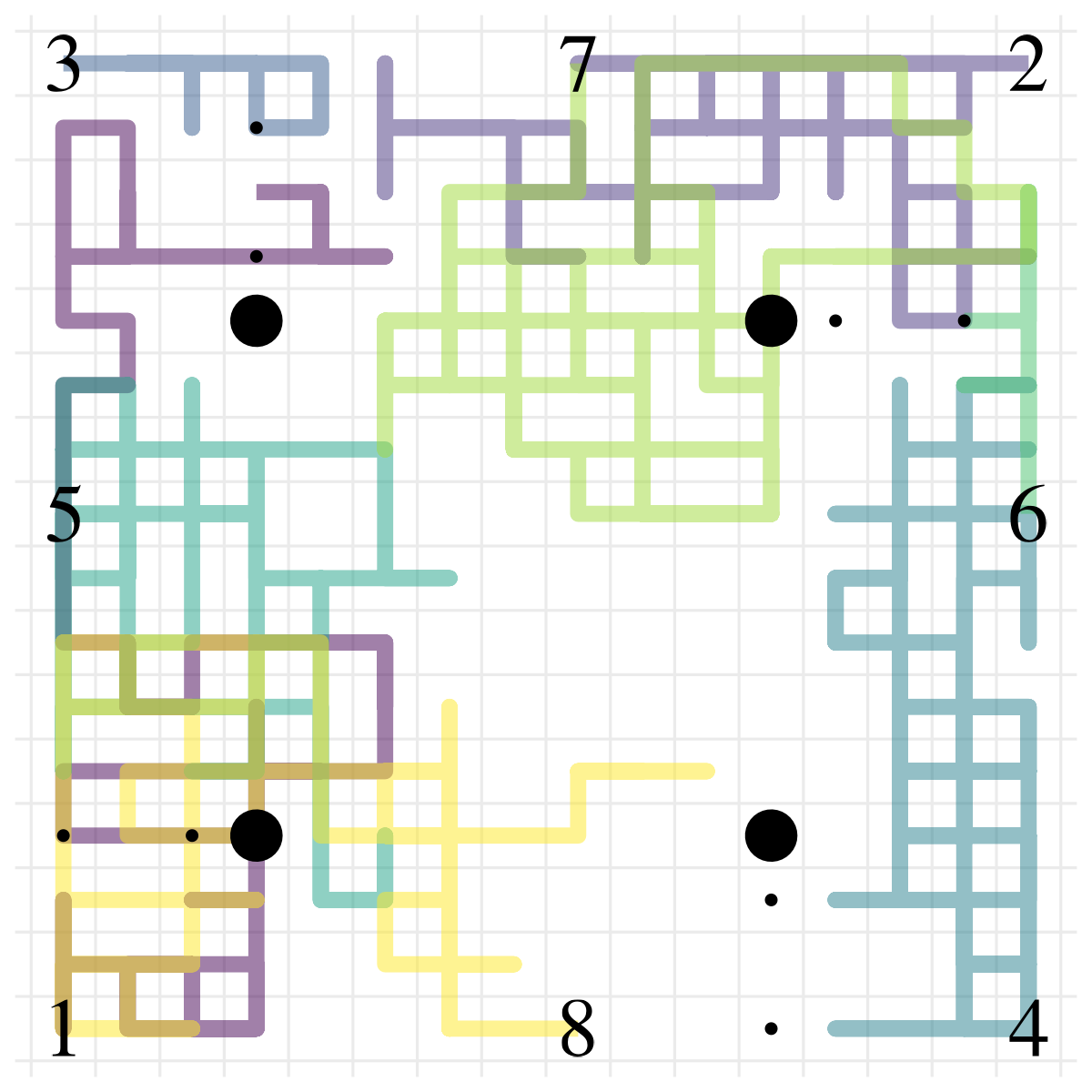}
    \caption{
        A Quandary Den illustration: Large circles indicate the fixed opponent positions and small circles indicate the direction of their projectiles; when the player character starts are `corners', they start from the numbered positions in order, and when they are `same' they all start at position 1; sample evolved paths shown by colored lines.
    }
    \label{fig:showQuandaryDen}
\end{figure}

If science boils down to physics and stamp collecting, artificial life affords a third option: making new stamps.
For this exploration of ``life as it could be,'' science fiction provides a head start from which possibilities can be drawn.
Hence, the Quandary Den, a game inspired by the Danger Room of the X-Men \citep{XMen63}.
It is also similar to the combat system of tabletop role-playing games.
The player wins a session by attacking opponent characters until the opponents' health is depleted, while also avoiding attacks; the player loses if all their characters' health is depleted or they exhaust all their actions.
The team score is the final health of the opponent characters less any health lost by the player's characters.

The game is played on an $n$ x $n$ grid with unconnected boundaries.
Actions are taken in discrete time steps, after which the outcome of those actions are resolved in the order they were initiated, based on character priority determined at the beginning of the session.
The selection of actions could come from human choices, dice rolls, algorithmic agents, or other inputs.
The available actions are also modifiable to accommodate a wide range of character types.
Here, three actions are used: moving one grid square, a melee attack to an adjacent square, or a ranged attack that travels in a straight line until it encounters a character or reaches the boundary.
These actions can be taken in one of the four cardinal directions.

In these experiments, the opponent locations are fixed, and they each employ a ranged attack in a single direction on a loop as illustrated in \Cref{fig:showQuandaryDen}.
Player characters take sequences of actions determined by genes.
Genes are encoded as text strings in the familiar series of As, Cs, Gs and Ts.
These are interpreted using a very simple grammar that only allows for expressions of token pairs: first a direction, then an action.
Each of the four letters are mapped to a direction, and two letters each are mapped to either `move' or `attack'.
The type of attack, melee or ranged, is set by the player character type.

Genes are decoded to action sequences and provided to the game engine at the start of a session.
A player character will attempt to execute a sequence like ``move left, move left, melee attack up, move down'' in order.
The outcome of ``melee attack up'' will depend on whether anyone is in the square above the character.
Multiple characters can occupy the same grid square without consequence.
Attacks take health from the first character they encounter, and the highest priority character if multiple characters are in the same square.

Since it features actors moving in a grid and carrying out actions to achieve an objective, the Quandary Den is akin to Robby the Robot \citep{MM1}.
The significant difference is that here, genes encode the specific actions of the player characters, rather than a strategy for choosing actions in response to the environment.
This is more similar to the model of \citet{HA1}, albeit with a very different fitness function.
Another difference from the models of \citet{MM1} and \citet{HA1} is the possibility for a single genome to include multiple genes representing the action sequences of separate characters.

\section{Details of Evolution and Implementation}

\begin{figure*}[!t]
    \centering
    \includegraphics{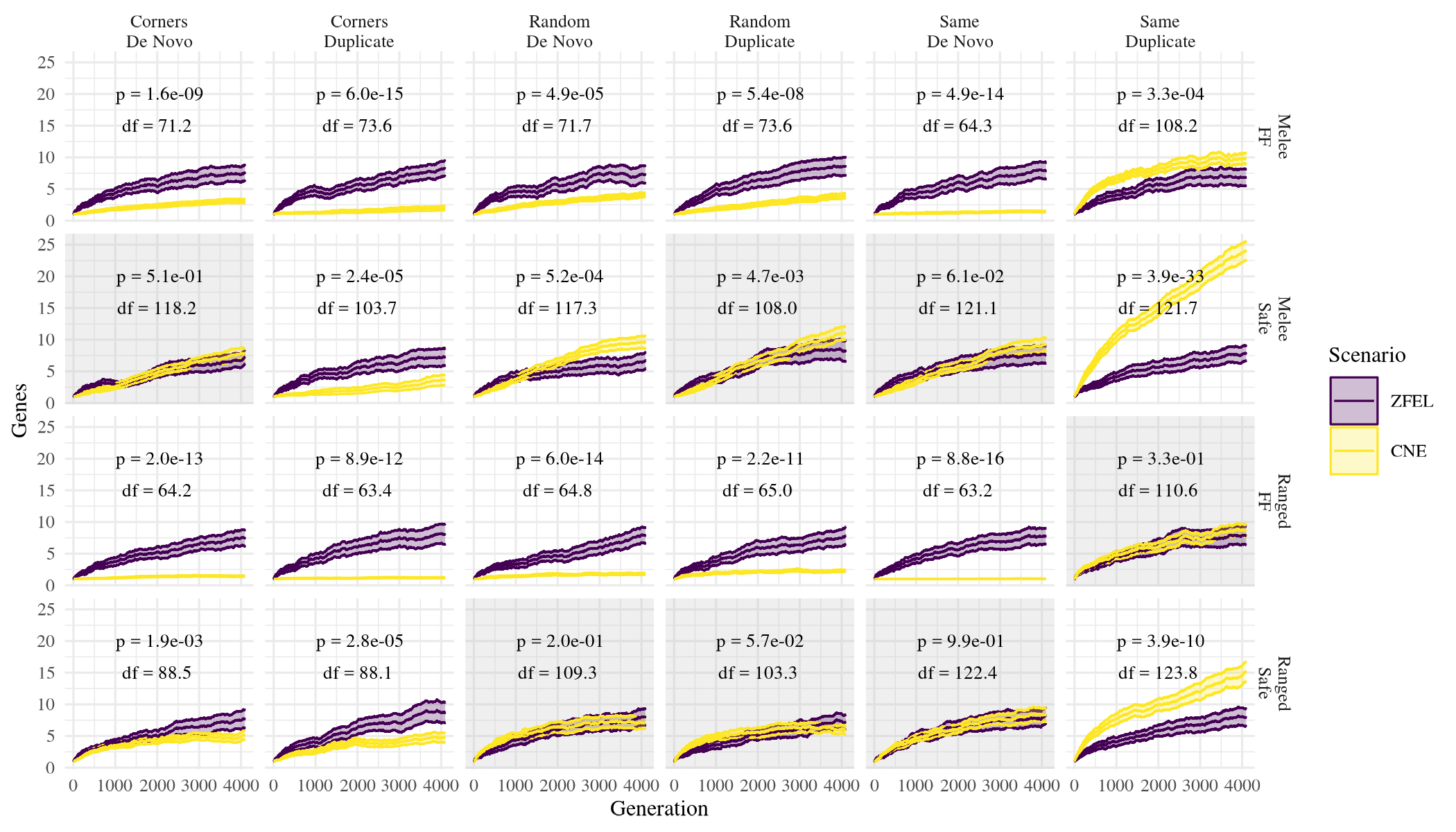}
    \caption{
        Number of total genes over time, with the ribbon indicating the 95\% confidence interval; player start scheme and gene origin method vary horizontally, and player character type and team safety vary vertically; the distributions at generation 4,096 were compared between the ZFEL and CNE scenarios by $t$ test, with the $p$ value and degrees of freedom printed per plot; shaded plots did not have a significant difference at p $<$ 0.05 with a Bonferroni correction for the 24 tests.
    }
    \label{fig:testLongerCNEGenes}
\end{figure*}

To evolve functional complexity neutrally, a functional solution is needed to start with.
All experiments begin with a haploid genome containing one gene with a random sequence 512 characters long.
Individuals are selected randomly from the existing population to reproduce with random point mutations at a fixed rate per site. 
The game score of the offspring is assessed, and if it is at least as high as the parent, the offspring is added to the next generation; otherwise it is discarded.
In essence, all score-lowering mutations are taken to be lethal.
Reproduction with mutation is repeated until the next generation has a number of individuals equal to the population size.
A population of one was used for initial experiments to minimize selection for second-order effects such as evolvability.
Additional generations are created in the same fashion until a genome achieving the maximum possible score is identified.
This genome becomes the starting point for the neutral evolution of complexity.

At that point, the possibility of adding or removing genes is introduced.
A gene can be added and/or removed; the probability of both events are kept equal to avoid bias in either direction and are assessed independently.
For removal, one of the available genes is chosen with uniform probability; if there is only one gene, nothing happens.
For addition, two different origins were tested separately.
Duplication involves selecting one of the existing genes with uniform probability and copying its sequence.
\emph{De novo} origin involves generating a new random sequence.
In both cases, the new gene is added to the end of the gene set, making it the lowest priority player character.

After the initial solution is found, the possibility of continuing to employ negative selection or not is introduced.
In the absence of negative selection, all generated offspring are viable for the next generation regardless of score.
In the presence of negative selection, all score-lowering mutations continue to be treated as effectively lethal.

In addition to testing player attack types (melee vs ranged), gene origin methods (duplication vs \emph{de novo}, and presence or absence of negative selection (CNE vs ZFEL), several other variables of the Quandary Den were explored.
Player characters can be subject to friendly fire attacks from their teammates (FF) or they can be kept \emph{safe} from such attacks.
Player characters can all start the session from the \emph{same} location, from \emph{random} locations assigned when the gene is created and then fixed, or from alternating \emph{corners} to minimize interference.

From the start with a random genome, an evolution trial proceeds for however many generations are needed to identify a solution, and then for a fixed number of generations afterwards.
The Quandary Den conditions are held constant throughout a given trial.
A set of 64 trials was carried out for each combination of conditions tested.

The game simulation and the evolutionary processes are implemented in Javascript.
Experiments were executed in R \citep{RSoftware} using the V8 package to run the Javascript code \citep{V8Software}.
Results were processed and analyzed in R \citep{TidySoftware} as well and visualized with the ggplot2 package \citep{ggplotSoftware}.
Code for running the experiments and analyzing the results, as well as the generated data, are available \citep{Walsh26}.

\begin{figure*}[!t]
    \centering
    \includegraphics{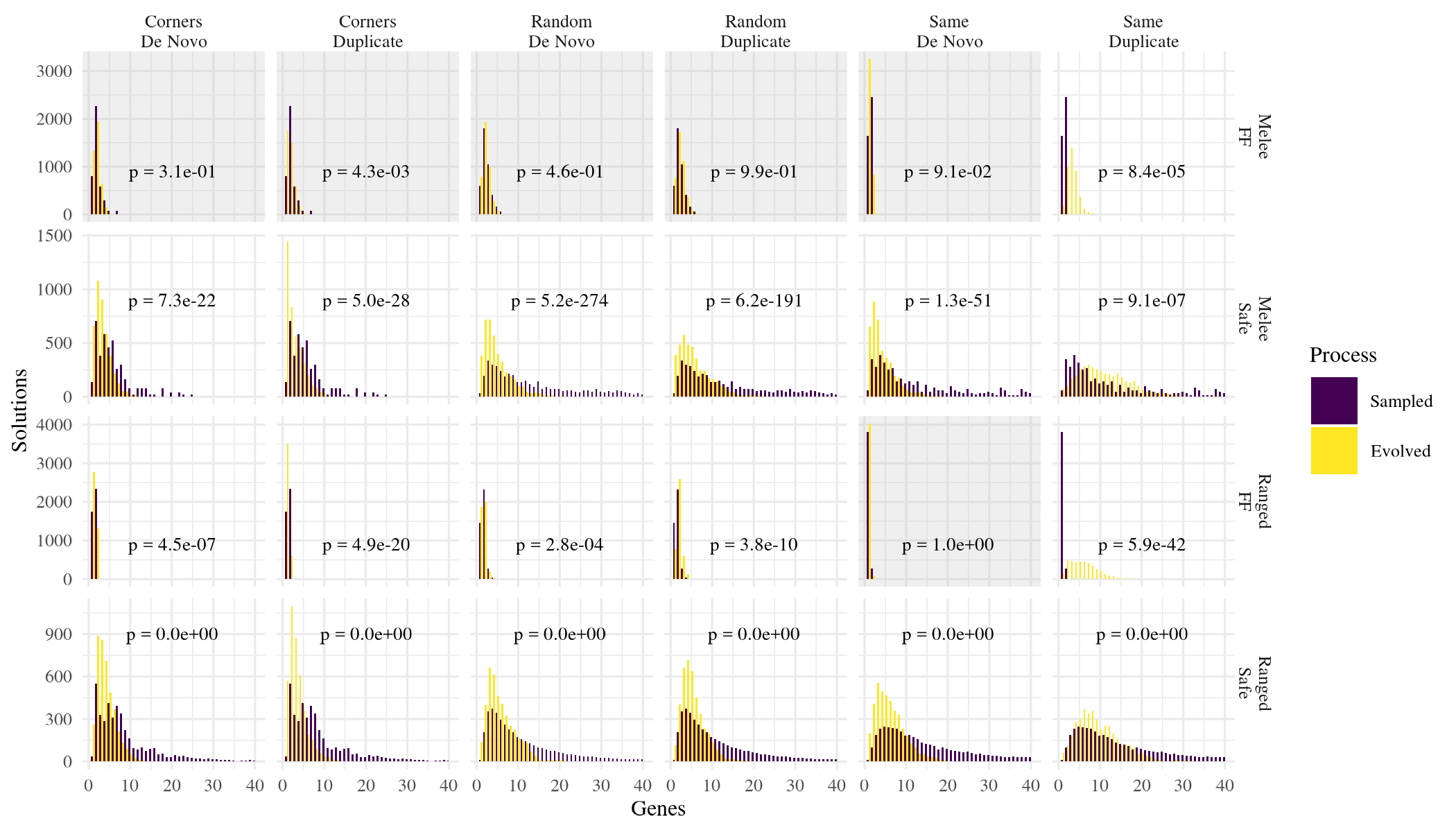}
    \caption{
        Number of solutions with the given number of genes from random sampling and evolution; values were adjusted to a common scale for visual comparison; player start scheme and gene origin method vary horizontally, and player character type and team safety vary vertically; the distributions from sampling and evolution were compared by two-sided K-S test, with the $p$ value printed per plot; shaded plots did not have a significant difference at p $<$ 0.05 with a Bonferroni correction for the 24 tests.
    }
    \label{fig:testGeneCountDist}
\end{figure*}

\section{Evolution Parameters}

When assessing the number of genes that accumulate, the rate at which they are added and removed will obviously be significant.
Therefore, evolution trials were done for rates 0.01, 0.005, 0.001, 0.0005, and 0.0001 and the number of genes compared from generations 100, 200, 1000, 2000, and 10000, respectively.
As expected, with no selection, the number of genes at those times were found to not be distinguishable by ANOVA, whether new genes were \emph{de novo} (F = 0.92, p = 0.45) or duplicates (F = 0.52, p = 0.72).
This was also true under selection, for both \emph{de novo} genes (F = 1.0, p = 0.41) and duplicates (F = 0.91, p = 0.46).
For subsequent experiments, for every reproduction there was a 1\% chance of adding a gene and an independent 1\% chance of removing a gene.

The number of genes may also be sensitive to the point mutation rate.
For example, it may be easier to remove genes at lower point mutation rates since there may not have been time for a nonessential gene to be rendered essential before it is picked to be removed.
Trials were carried out with rates of 0.05, 0.01, 0.005, 0.001, and 0.0005 and the number of genes compared after $2^10$ generations.
With no selection, the number of genes did not vary by rate, as determined by ANOVA, when new genes were duplicated (F = 0.63, p = 0.64).
When new genes were \emph{de novo}, ANOVA analysis indicated variability by rate (F = 3.8, p = 0.0053), but there was no apparent trend with rate.
With selection, the number of \emph{de novo} genes did not vary with rate (F = 0.50, p = 0.73) but did change significantly for duplicate genes (F = 26, p = 2.7 x $10^{-18}$).
The fewest genes accumulated with the rate of 0.05, and comparable numbers accumulated for the rates 0.005, 0.001 and 0.0005.
Therefore, for subsequence experiments, a point mutation rate of 1\% per site per generation was used, as this yielded intermediate results with selection, and was not qualitatively inferior to any other option without selection.

\section{Changes in Complexity}

\begin{figure*}[!t]
    \centering
    \includegraphics{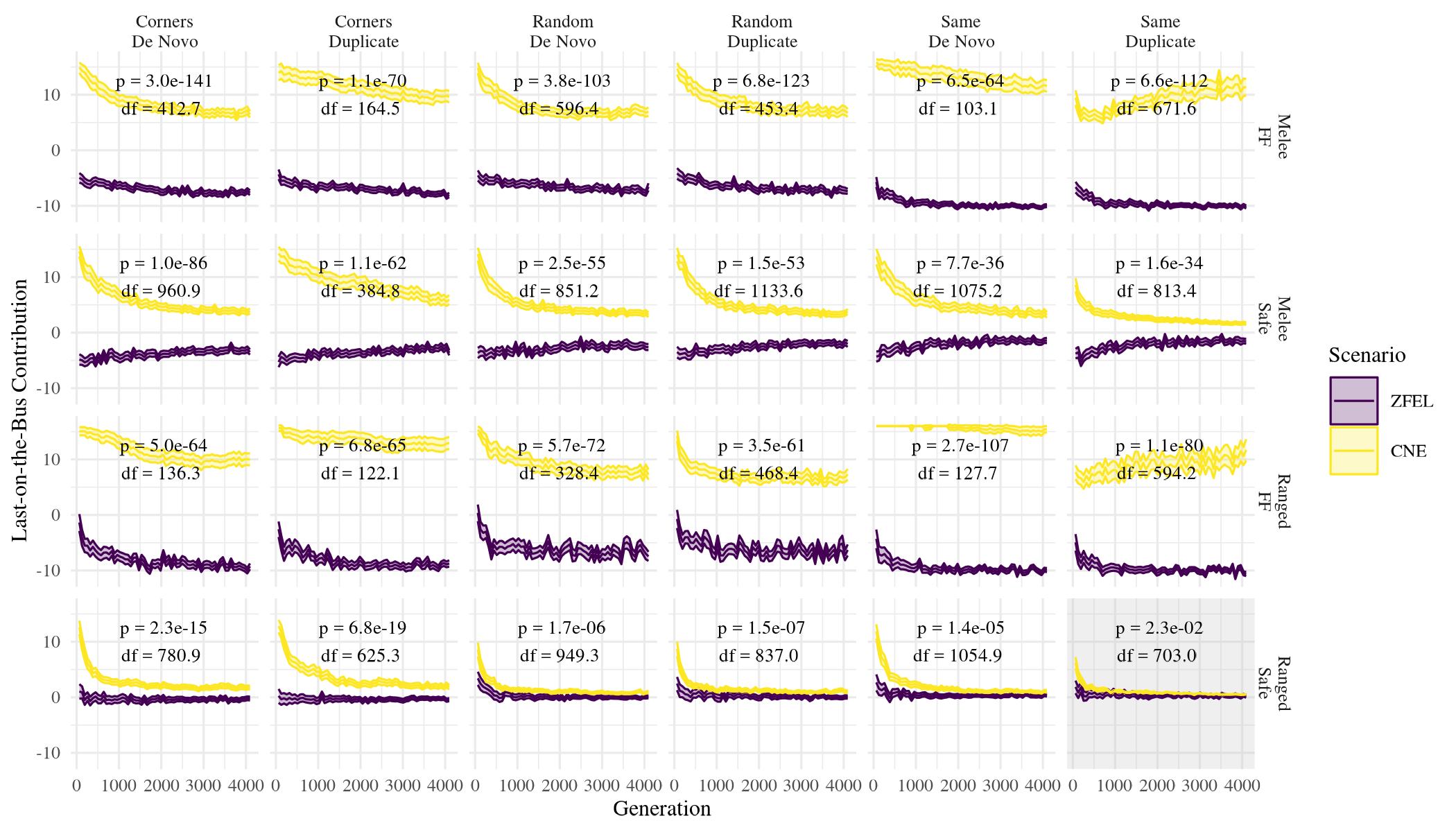}
    \caption{
        Mean LotB contribution per gene over time, with the ribbon indicating the 95\% confidence interval; player start scheme and gene origin method vary horizontally, and player character type and team safety vary vertically; the distributions at generation 4,096 were compared between the ZFEL and CNE scenarios by $t$ test, with the $p$ value and degrees of freedom printed per plot; shaded plots did not have a significant difference at p $<$ 0.05 with a Bonferroni correction for the 24 tests.
    }
    \label{fig:testLongerCNELotB}
\end{figure*}

\citet{McShea19} use number of distinct parts as one simple measure of complexity.
By that metric, ZFEL and CNE scenarios were compared, with all three possible results obtained depending on game conditions (\Cref{fig:testLongerCNEGenes}).
All combinations of the following parameters were tested: \emph{de novo} new genes and duplicated new genes, player characters all starting in the `same' place and spread out in separate `corners' and in `random' places, melee and ranged attacks, and keeping player characters `safe' from their teammates' attacks and subjecting them to friendly fire (`FF').
In most cases, typically with friendly fire, CNE led to fewer genes accumulating compared to ZFEL.
In some cases (shaded grey charts), the distribution of genes at generation 4,096 could not be distinguished.
In a few cases, CNE led to more genes, most dramatically for melee characters starting from the same position with (initially) duplicated genes and safeties from teammates.

As expected, in all cases the ZFEL scenario led to the same distributions of gene number over time.
For the CNE scenario, there were substantial differences.
The different parameters of the Quandary Den conditions have the potential to significantly change the geometry and topology of the space of maximum score solutions.
This creates more opportunity to assess whether the gene-size of solutions can be explained by ergodic exploration of solution space or if there is a ratcheting effect towards higher gene numbers with corresponding smaller partitions of that space.

Randomly drawing from the space of maximum score solutions is challenging.
As an alternative, sequences were randomly sampled for different numbers of genes and maximum score solutions counted from $2^16$ samples per gene number and game conditions.
For the melee attack characters, this number of samples was not sufficient to find a sample of maximum score solutions with sufficient statistical power.
To adjust for this, evolution trials were repeated that only required at least half the maximum score to be viable for the next generation.
These results were compared against the randomly sampled solutions that achieved at least half the maximum score.

Distributions of number of genes among the half and full score solutions from the random sampling and evolution trials are compared in \Cref{fig:testGeneCountDist}.
For visualization purposes, the numbers are rescaled by calculating rates and multiplying by a common number of samples.
Unadjusted samples were used for the two-sided Kolmogorov-Smirnov test to assess whether the two samples were drawn from the same distribution \citep{SCHROER1995185}.
In most cases (charts with white backgrounds), the distributions were different at a p $<$ 0.05 level with a Bonferroni correction for the 24 tests.
When the evolved distribution is skewed to smaller numbers relative to the randomly sampled distribution, there is the possibility that this difference would disappear with more generations; therefore, no strong conclusions should be drawn from these observations.
But the FF conditions for both melee and ranged characters starting at the same positions with initially duplicated genes show skewing towards higher gene numbers, consistent with ratcheting.

\section{Consequences of Ratcheting}

\begin{figure}
    \centering
    \includegraphics[width=.4\textwidth]{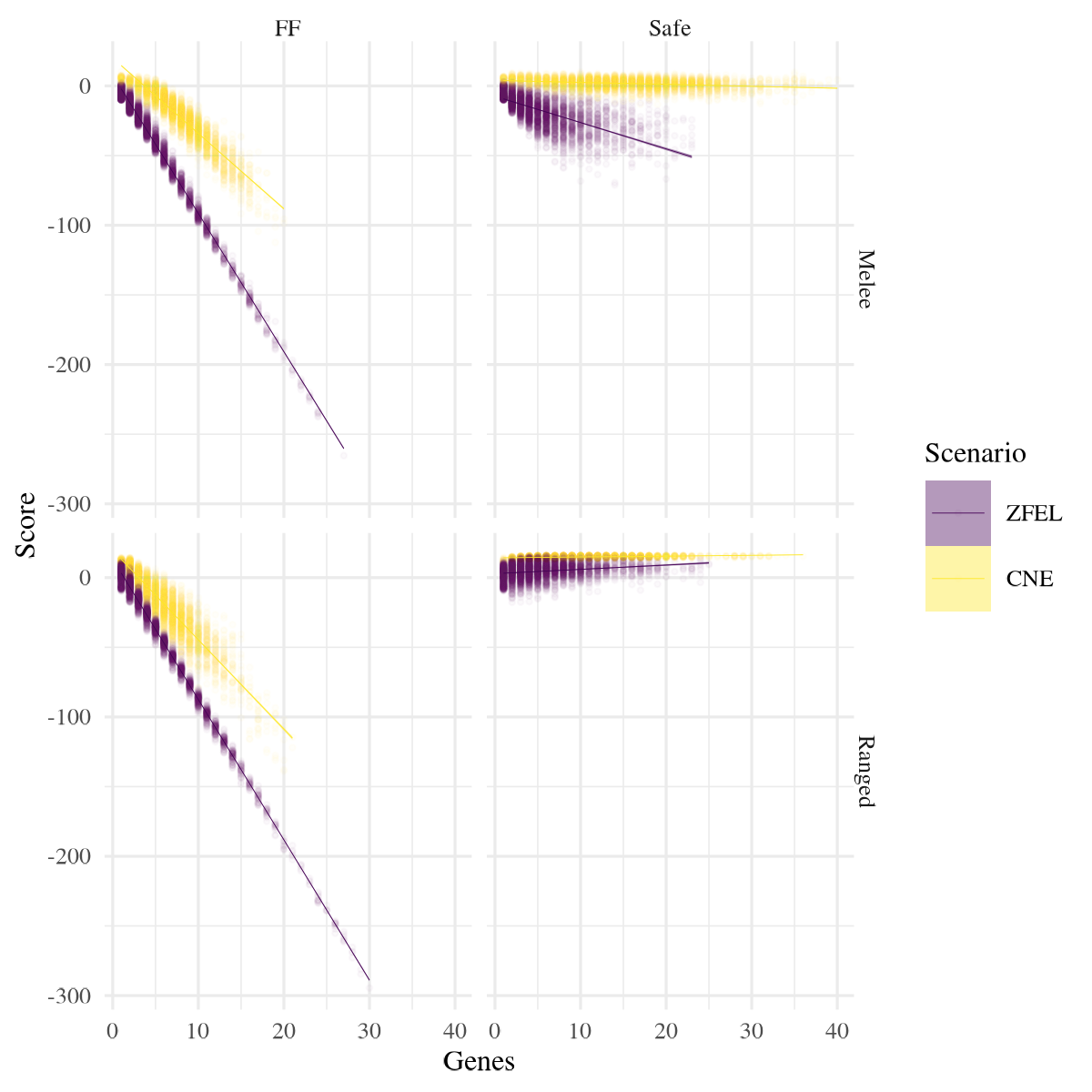}
    \caption{
        Robustness as it varied by total gene number, with a linear model fit to each scenario; team safety varies horizontally and player character type varies vertically; only genes created by duplication and `same' player start scheme shown.
    }
    \label{fig:testRobustnessByGenes}    
\end{figure}

Gene number is a fairly crude measure of complexity.
Adding more and more copies of the same gene can, in some of these Quandary Den conditions, have little or no impact and represent redundancy rather than complexity.
More significant to complexity is increasing the number of essential genes.
The Last on the Bus (LotB) contribution can be used to identify essential genes \citep{Page18}.
For a given gene/character, it is the difference between the score of the entire team and the score of the team minus that character.

\begin{figure*}
  \centering
  \includegraphics{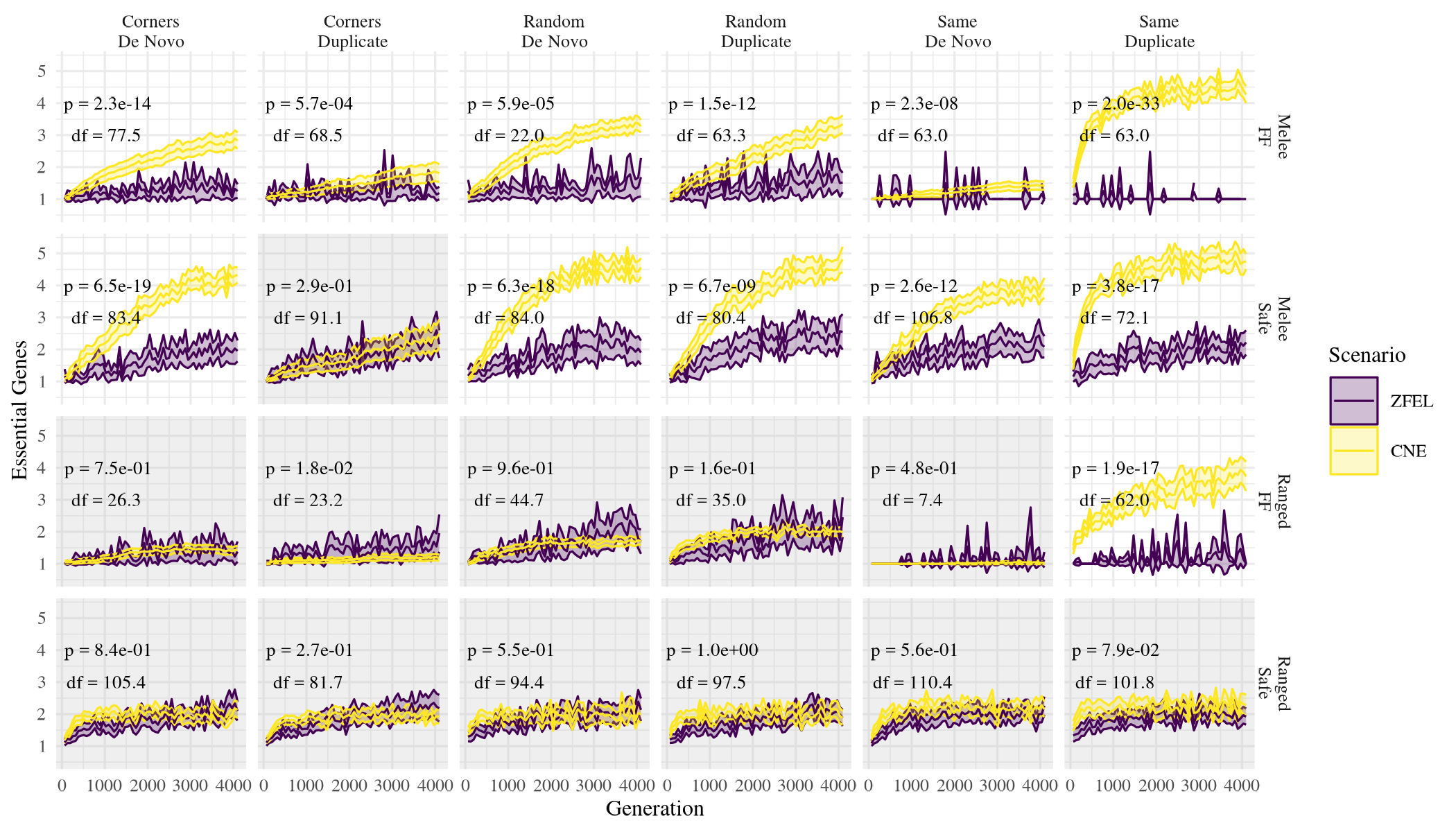}
  \caption{
    Number of essential genes over time, with the ribbon indicating the 95\% confidence interval; player start scheme and gene origin method vary horizontally, and player character type and team safety vary vertically; the distributions at generation 4,096 were compared between the ZFEL and CNE scenarios by $t$ test, with the $p$ value and degrees of freedom printed per plot; shaded plots did not have a significant difference at p $<$ 0.05 with a Bonferroni correction for the 24 tests.
  }
  \label{fig:testEssentialGeneDifference}
\end{figure*}

The number of essential genes, meaning those with a positive LotB contribution, also increases over time under most conditions.
It actually varies between conditions in the ZFEL scenario, and it is never smaller under the CNE scenario than the ZFEL scenario for a given set of conditions, although in all but one of the ranged attack conditions the two scenarios produce the same number of essential genes (\Cref{fig:testEssentialGeneDifference}).

Under most conditions, the average LotB contribution declines over time as the number of genes increases (\Cref{fig:testLongerCNELotB}).
This can be thought of as a form of subfunctionalization, with different characters contributing some of the necessary attacks.
For the CNE scenario, the limit for this seems to be the average contribution of the random genes from the ZFEL scenario; it is difficult to see how it could get lower.
At this point, there is interlocking complexity, but it is loosely coupled; in general, any given gene could be replace by a new random gene.

For both melee and ranged attacks when the characters all start in the same position, friendly fire is active and genes are initially duplicated, something qualitatively different from sub-functionalization occurs.
The average contribution declines at first, then increases.
In some instances, it gets substantially higher than the maximum score.
What is happening when the LotB contribution increases is that the removal of one gene/character results in the Quandary Den exercise continuing longer than when that character is present.
As a result, the remaining players wind up landing attacks on each other and taking away health, lowering the score potentially below zero.
These attacks can't be selected against when the exercise ends at the earlier point.

As new genes get added, there is the potential for the session to end even earlier.
This, plus the addition of another character, creates even more opportunities for these masked friendly fire attacks.
This creates the opportunity for ratcheting beyond the most likely number of genes.
It becomes nearly impossible to remove an added gene because removing it will unmask those attacks.

\begin{figure*}
  \centering
  \includegraphics{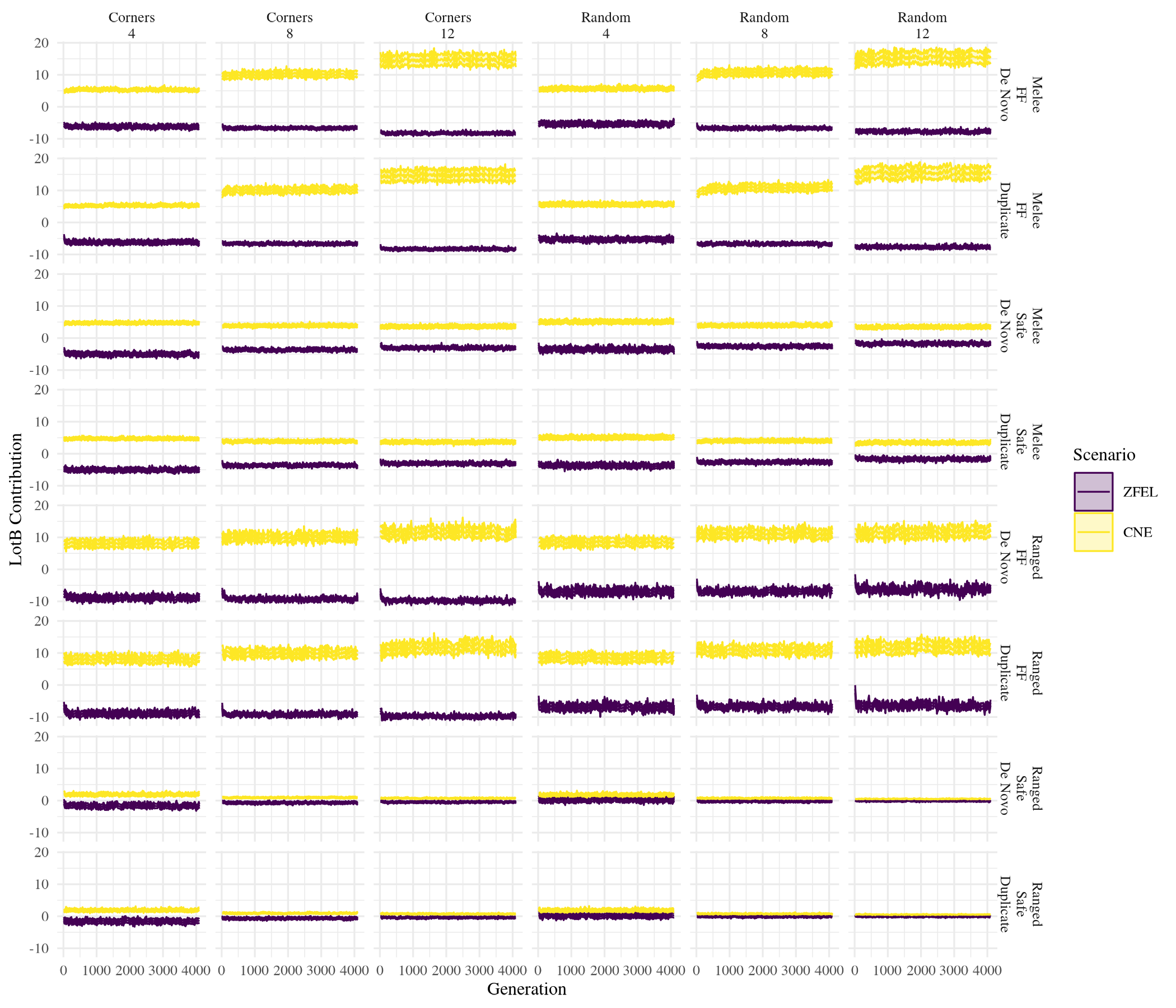}
  \caption{
    Last-on-the-Bus Contribution over time when gene number is fixed, with the ribbon indicating the 95\% confidence interval; player start scheme and number of genes vary horizontally, and player character type, team safety and gene origin method vary vertically.
  }
  \label{fig:testFixedGeneNumber}
\end{figure*}

In principle, the effect on LotB contribution should apply to the other start positions and other gene origin methods, but may not have been observed because the necessary number of genes were not reached in the allotted generations.
To test this, evolution trials for those game conditions were carried out with fixed numbers of genes---4, 8, and 12.
While LotB contribution did not change over generations, it did increase with increasing gene number under friendly fire conditions as expected; see \Cref{fig:testFixedGeneNumber}.

\begin{figure}
  \centering
  \includegraphics[width=.4\textwidth]{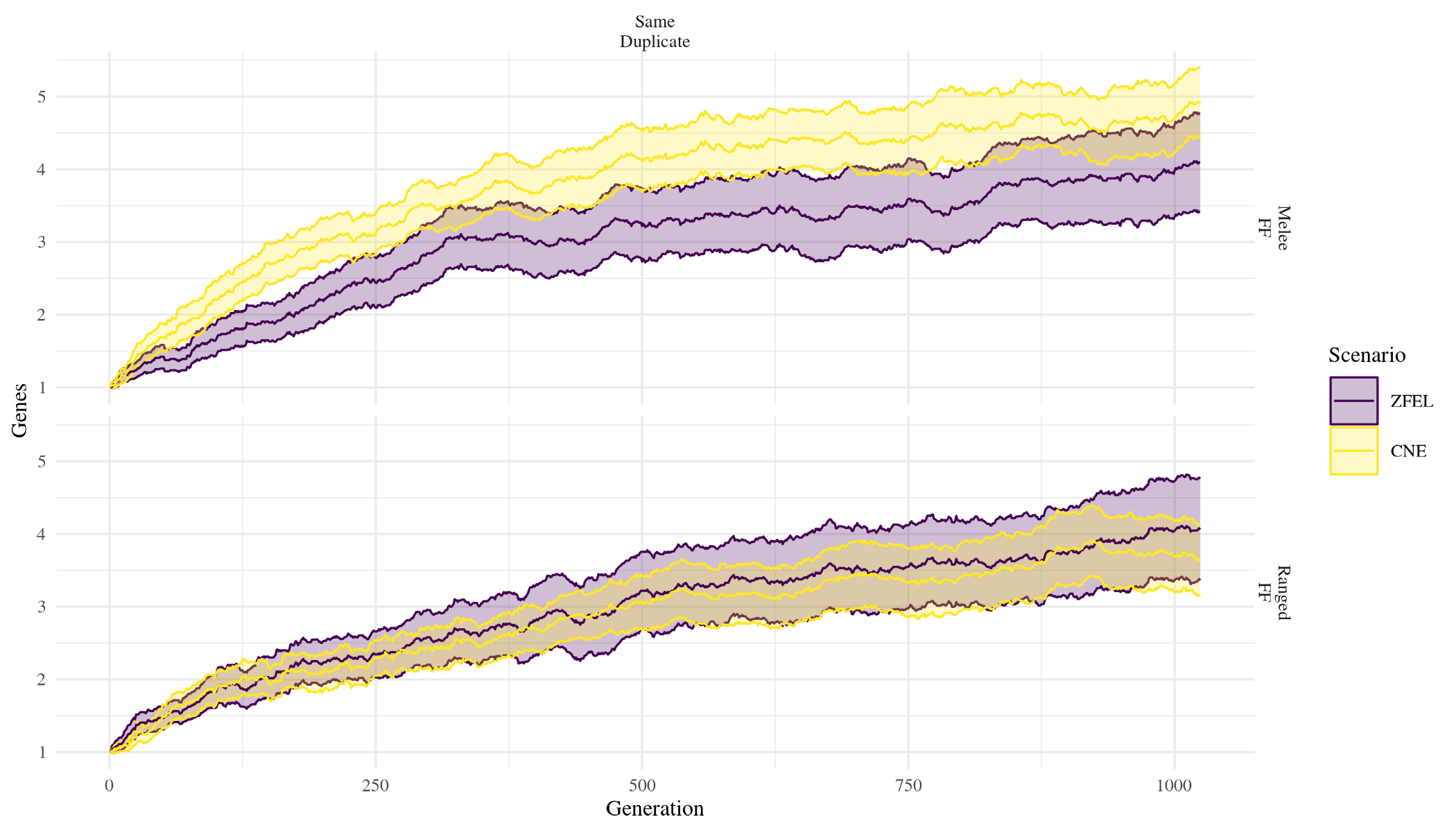}
  \caption{
    Number of total genes over time for random opponent configurations, with the ribbon indicating the 95\% confidence interval; player character type varies vertically.
  }
  \label{fig:testRandomOpponentsTotalGenes}
\end{figure}

\begin{figure}
  \centering
  \includegraphics[width=.4\textwidth]{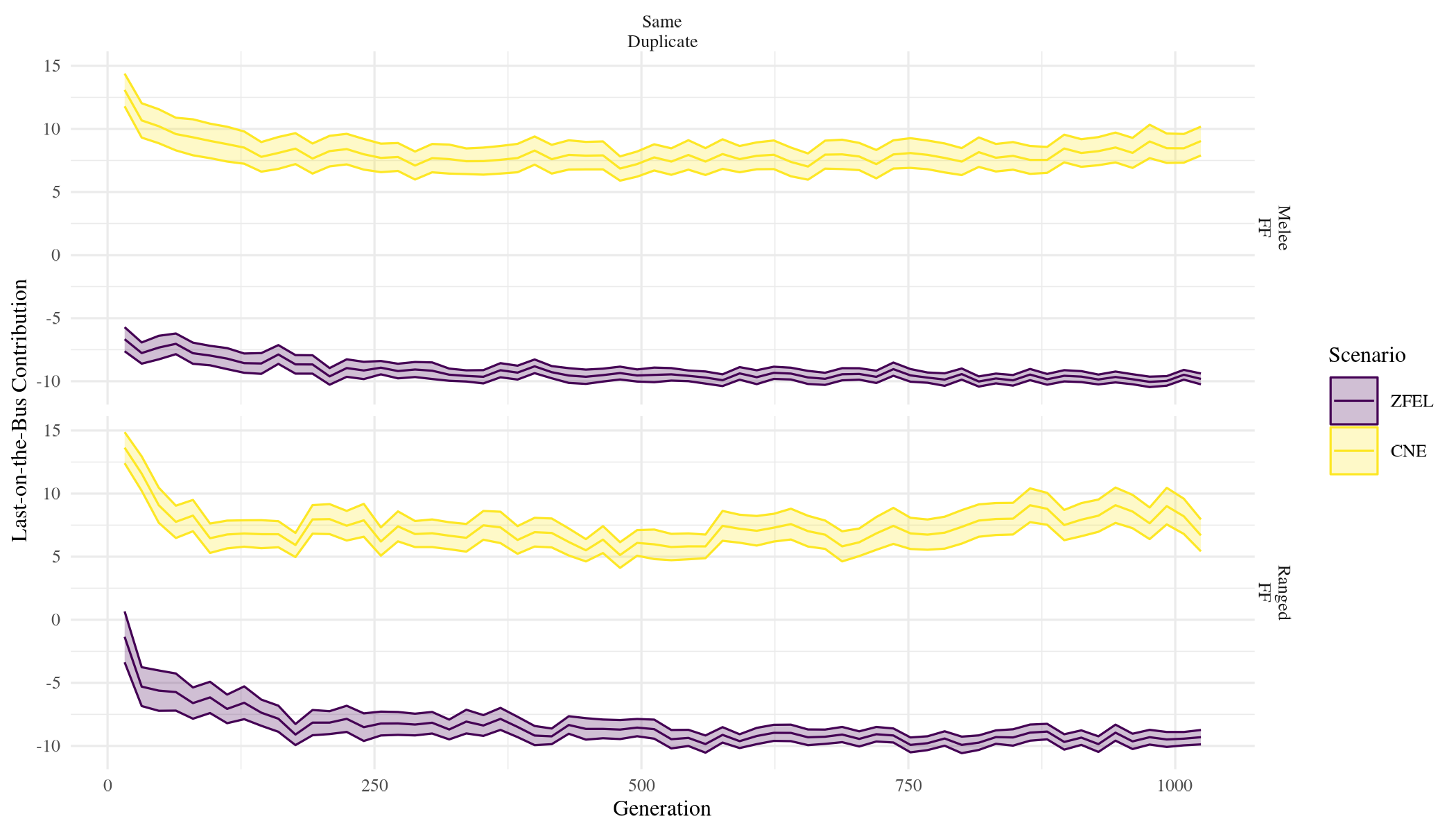}
  \caption{
    Last-on-the-Bus Contribution over time for random opponent configurations, with the ribbon indicating the 95\% confidence interval; player character type varies vertically.
  }
  \label{fig:testRandomOpponentsLotB}
\end{figure}

\begin{figure}
  \centering
  \includegraphics[width=.4\textwidth]{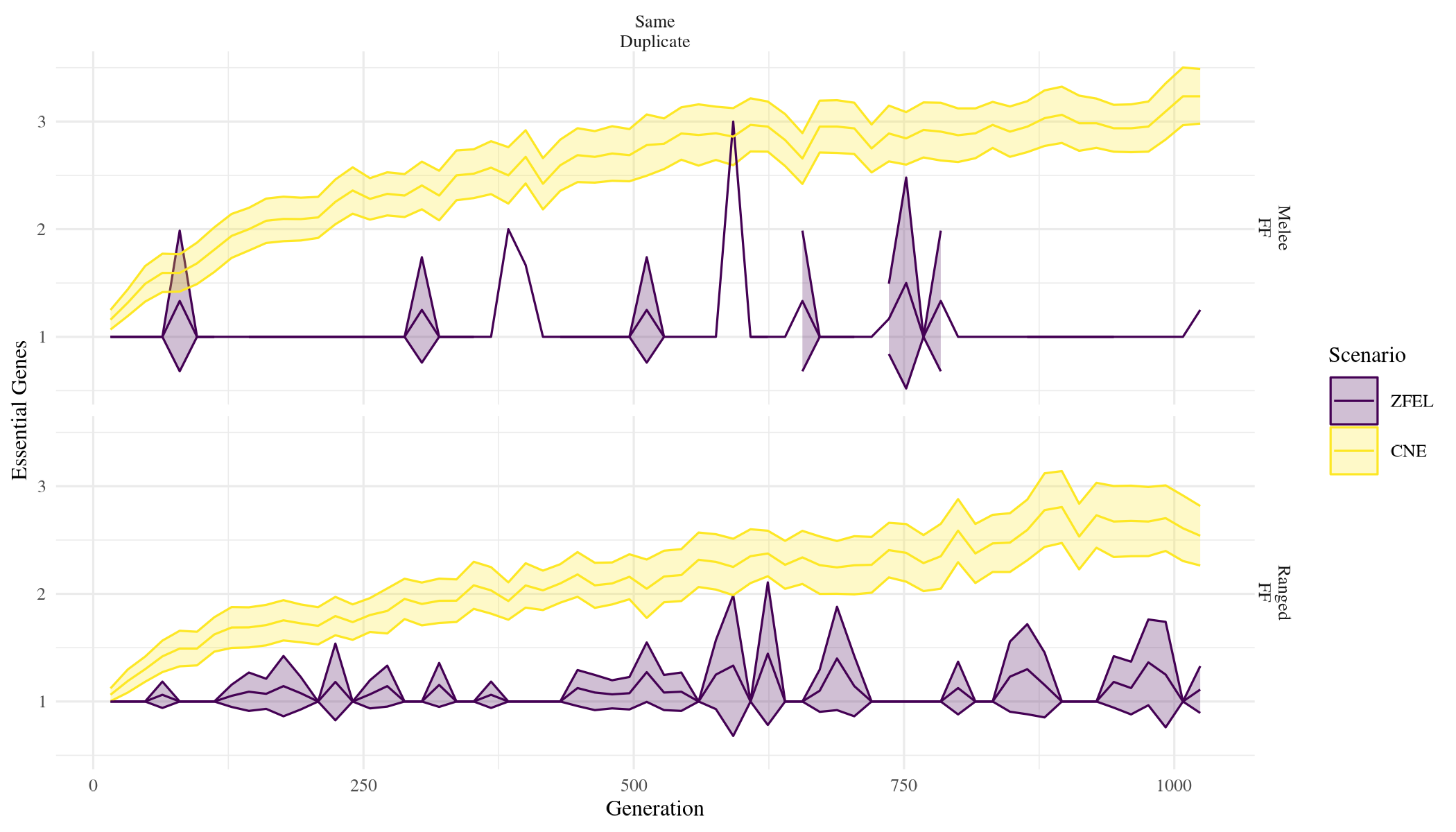}
  \caption{
    Number of essential genes over time for random opponent configurations, with the ribbon indicating the 95\% confidence interval; player character type varies vertically.
  }
  \label{fig:testRandomOpponentsEssentialGenes}
\end{figure}

To rule out the possibility that the ratcheting is an artifact of the specific opponent configuration, 64 trials with distinct random configurations were run.
After 1,024 generations, the results for gene number, LotB contribution and essential gene number were consistent with the results using the configuration shown in \Cref{fig:showQuandaryDen}; see \Cref{fig:testRandomOpponentsTotalGenes,fig:testRandomOpponentsLotB,fig:testRandomOpponentsEssentialGenes} for these results.

\begin{figure*}
  \centering
  \includegraphics{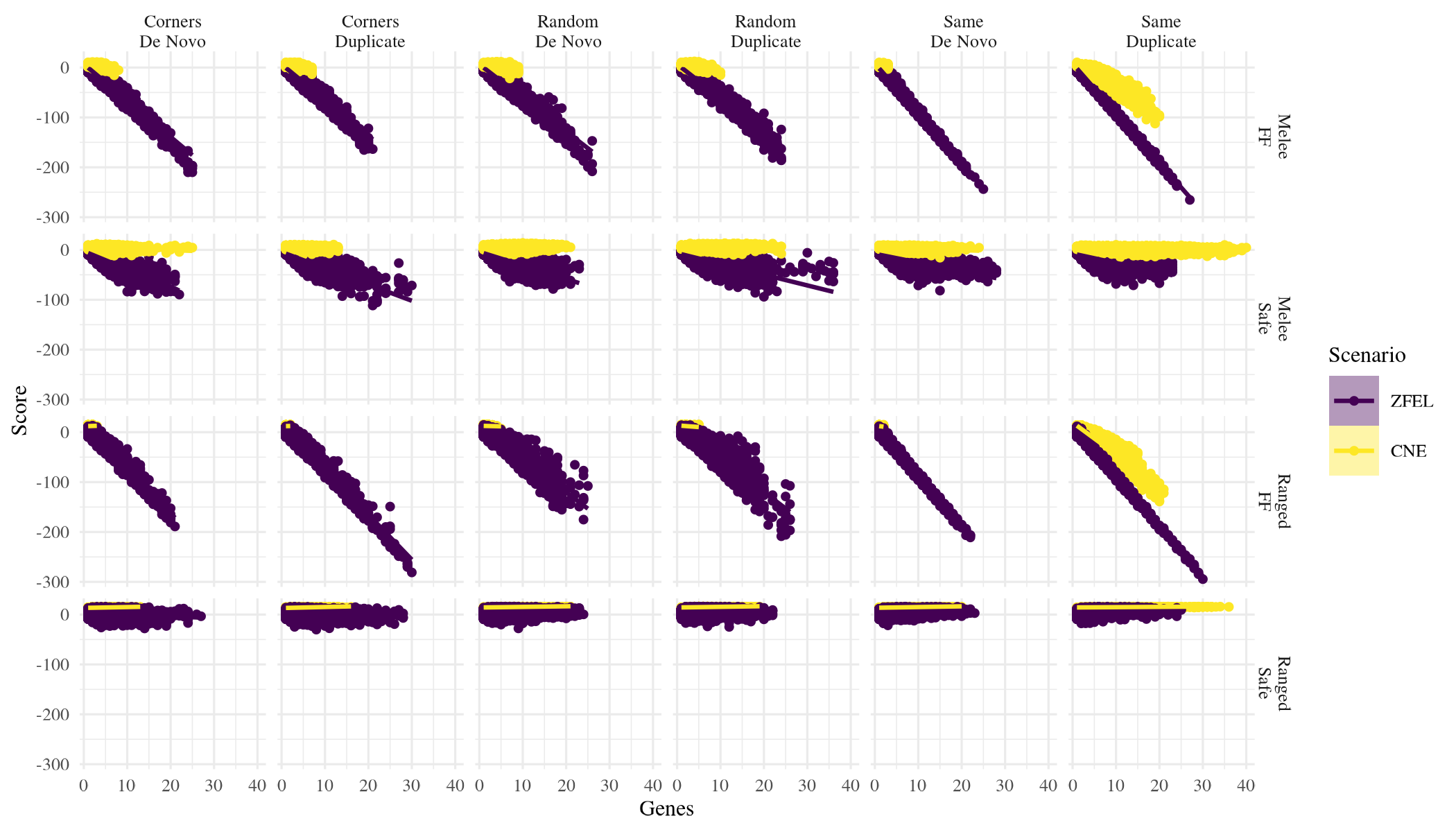}
  \caption{
    Robustness as it varied by total gene number, with a linear model fit to each scenario; player start scheme and gene origin method vary horizontally, and player character type and team safety vary vertically.
  }
  \label{fig:testRobustnessByGenesFull}
\end{figure*}

Since the number of solutions decreases with additional genes under the conditions where ratcheting was observed, the robustness of the solution is also impacted.
Robustness was assessed by subjecting a given solution to a single step of point mutation at the same rate as elsewhere and measuring game score.
This was repeated 64 times per individual and the results averaged.
\Cref{fig:testRobustnessByGenes} shows how this robustness declines with each additional gene for select conditions; see \Cref{fig:testRobustnessByGenesFull} for full results.
This is true for both CNE and ZFEL scenarios, although the selection in the CNE scenario mitigates this somewhat, likely just because the starting point is higher.
This decline in robustness had an apparent consequence on the experiments.
At times, a given trial would get stuck, unable to find a next generation that maintained the maximum score and thus unable to complete the specified number of generations in a practical amount of time.
These trials had to be restarted, potentially skewing the results away from even more extreme outcomes.

Under some of the conditions, the fact that the number of possible solutions decreases with gene number means that eventually there may be no way to add a new gene without lowering fitness, limiting how far the ratchet can go and how much robustness can be impacted.
However, when genes are duplicated and the starting point is the same, there is always a way to add one more gene.
This is because the added character will start out moving and attacking in the same pattern as another character.
Since selection prevented that character from harming any teammates, the newly added character won't cause harm either.
But it does have the potential to end the session sooner, furthering the ratchet.

Robustness can also be thought of in terms of loss of whole genes as well as resilience to point mutations.
Since not all of the genes were found to be essential through the LotB contribution analysis, the multiple characters do provide some resilience against loss of genes.
And as the LotB contribution of individual genes declines, even essential genes can be more easily replaced, although this may be mitigated by the challenges of not interfering.

\begin{figure}
    \centering
    \includegraphics[width=.4\textwidth]{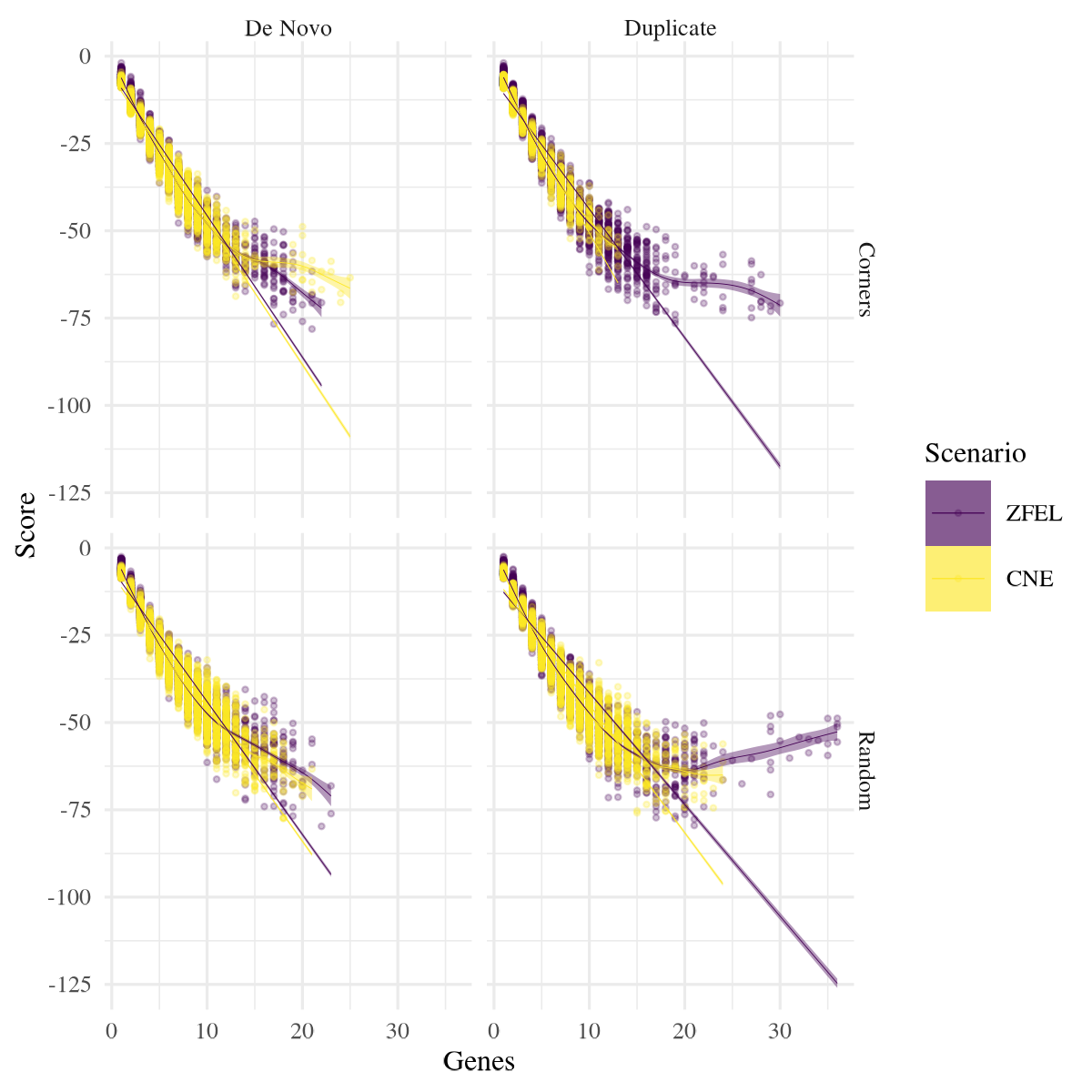}
    \caption{
         Plasticity as it varied by total gene number, with both a linear model and a nonparametric spline model fit to each scenario; gene origin method varies horizontally and player start scheme varies vertically; only melee player type and no friendly fire shown.
    }
    \label{fig:testPlasticityByGenes}    
\end{figure}

\begin{figure*}
  \centering
  \includegraphics{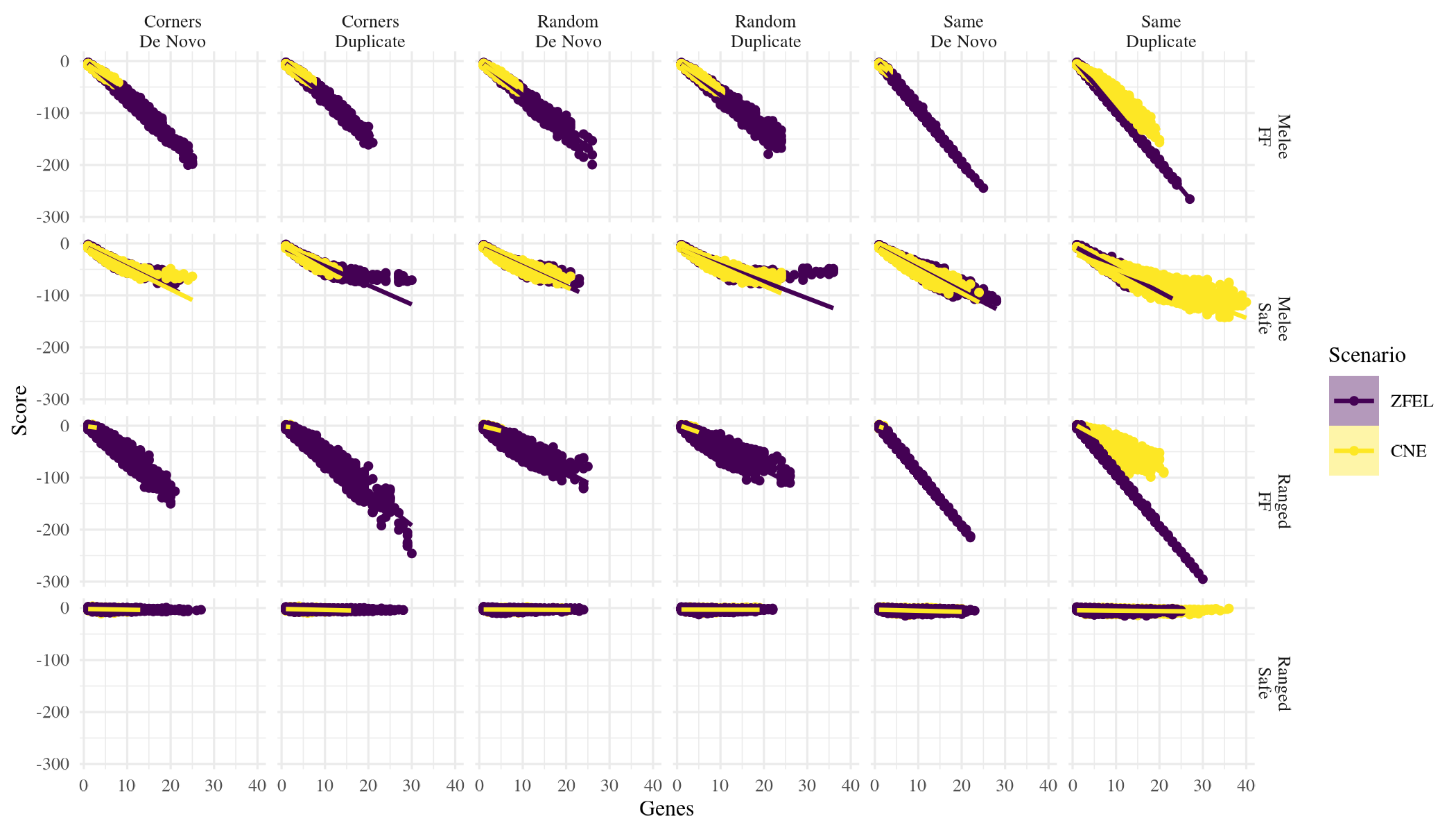}
  \caption{
    Plasticity as it varied by total gene number, with a linear model fit to each scenario; player start scheme and gene origin method vary horizontally, and player character type and team safety vary vertically.
  }
  \label{fig:testPlasticityByGenesFull}
\end{figure*}

Complexity could also potentially impact plasticity and evolvability.
To assess plasticity, 32 random opponent configurations were generated and the average score across all of them was measured.
In general, plasticity was seen to decline with increasing number of genes.
However, under some conditions, an inflection point was observed after which plasticity increased with the addition of more genes, as seen in \Cref{fig:testPlasticityByGenes}; see \Cref{fig:testPlasticityByGenesFull} for full results.
These conditions represent the least opportunity for interference, creating more of an opportunity for serendipitous synergy as more characters are present.

\begin{figure*}
  \centering
  \includegraphics{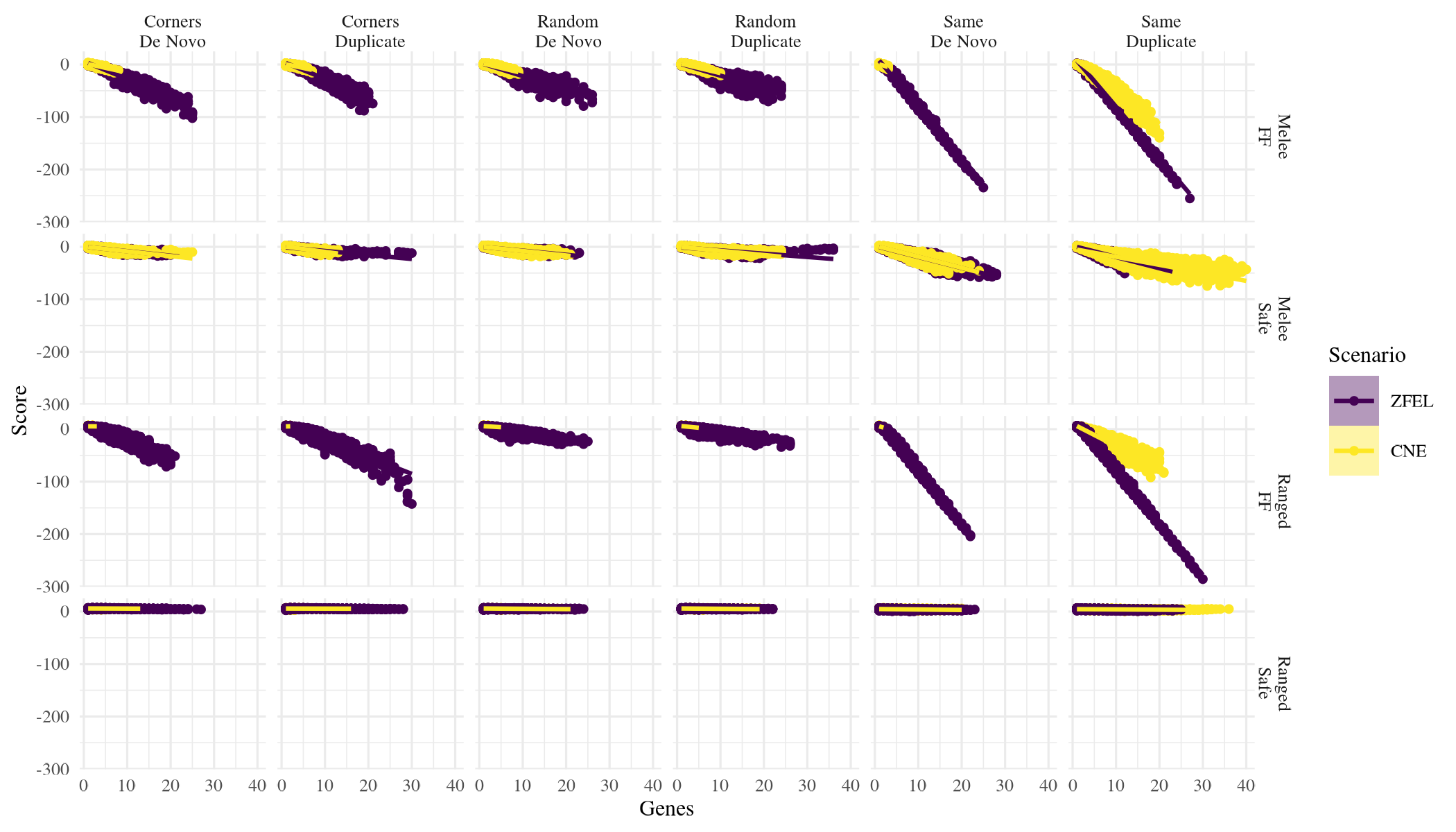}
  \caption{
    Evolvability as it varied by total gene number, with a linear model fit to each scenario; player start scheme and gene origin method vary horizontally, and player character type and team safety vary vertically.
  }
  \label{fig:testEvolvabilityByGenesFull}
\end{figure*}

To assess evolvability, 32 rounds of reproduction with mutation and gene addition \& removal were completed for 64 generations, and then the score on the random opponent configurations was measured.
No improvements in evolvability were observed with increasing numbers of genes for any conditions (\Cref{fig:testEvolvabilityByGenesFull}).

\section{Discussion}

While nominally the player characters of the Quandary Den are inspired by (fictional) agents with human cognition, their capabilities have little to do with sensing, reasoning individuals.
With a 3D (two sapce plus one time) structure determined by sequence and environment, and function mediated by specific ``residues'' at specific locations in that structure, what is evolving here is more comparable to proteins than people; more artificial molecular biology and biochemistry than cognitive psychology or neuroscience.
A one-to-one correspondence to proteins is not intended or achieved, and generalization should therefore be undertaken with appropriate reservation, but generalization to humans is even less reasonable.

With that in mind, it is notable that this model system replicates two kinds of interlocking complexity, both encountered in biological systems.
Subfunctionalization is seen in the complexity generated in the experiments by \citet{Depres24} and observed in the complexification of hemoglobin analyzed by \citet{Pillai20}.
By contrast, systems that require RNA editing like those discussed by \citet{Lukes11} and \citet{MunozGomez21} with respect to CNE representing a different scenario, where something that impairs function has to be removed.
This is akin to what was observed when the LotB contribution increased.

The robustness results in those scenarios suggest a potential hypothesis that could be applied to biological systems.
Perhaps some biological complexes have been ratcheted to a point where little further variability is possible.
For example, it is noted that there are substantial differences between bacterial and eukaryotic ribosomes, but within those domains ribosome features are conserved.
The lack of greater variety suggests substantial restrictions on possible realizations of ribosome functionality, raising the question of how a transition occurred at all.
Perhaps endosymbiosis or some other event early in eukaryote origins changed the landscape sufficiently to allow changes, followed by ratcheting to a relatively small portion of solution space.

Devising a way to test that hypothesis may take time.
In the interim, more can be explored in the Quandary Den.
In particular, larger population sizes could be tested to see how competition within generations might introduce further forms of selection.
Changing environments, rather than the status opponent configurations used here, may also add selective effects.

\section{Acknowledgements}

Although not directly funded or endorsed, this work was made possible by employment at Health Monitoring. Thanks to S. Joshua Swamidass for asking the question that led to the initial creation of the Quandary Den. Thanks to Matt Fraction, Simon Spurrier, Al Ewing, Jonathan Hickman, Ryan North, and especially Kieron Gillen for inspiration for this work.

\footnotesize
\bibliographystyle{apalike}
\bibliography{NeutralEvoInterlockingQuandaryDen}

\end{document}